\documentclass[journal]{IEEEtran}
\IEEEoverridecommandlockouts
\usepackage{amsmath,amssymb,amsfonts}
\usepackage{algorithmic}
\usepackage{graphicx}
\usepackage{textcomp}
\usepackage[table]{xcolor}
\usepackage{mathrsfs}
\usepackage{stmaryrd}
\SetSymbolFont{stmry}{bold}{U}{stmry}{m}{n}
\usepackage{bm}
\usepackage[caption=false]{subfig}
\usepackage{centernot}
\usepackage{pgf, tikz}
\usepackage{algorithm}
\usepackage{multirow}
\usepackage{hyperref}
\usepackage{soul}

\usetikzlibrary{matrix}
\usetikzlibrary{backgrounds}
\usetikzlibrary{calc}
\usetikzlibrary{shapes,positioning,angles,quotes}
   
\usepackage{tikz-cd}
\usepackage{pgfplots}
\pgfplotsset{compat=1.18}
\def\BibTeX{{\rm B\kern-.05em{\sc i\kern-.025em b}\kern-.08em
    T\kern-.1667em\lower.7ex\hbox{E}\kern-.125emX}}

\DeclareMathOperator*{\argmin}{arg\,min} 
\DeclareMathOperator*{\argmax}{arg\,max}

\let\oldpagenumbering\pagenumbering
\renewcommand{\pagenumbering}[1]{%
	\cleardoublepage
	\oldpagenumbering{#1}
}

\newcommand{\diff}{\mathop{}\mathopen{}\mathrm{d}}

\newcommand{\eu}{\ensuremath{\mathrm{e}}}

\providecommand{\renewoperator}[3]{%
	\renewcommand*{#1}{\mathop{#2}#3}}
\renewoperator{\Re}%
	{\mathrm{Re}}{\nolimits}
\renewoperator{\Im}%
	{\mathrm{Im}}{\nolimits}

\newcommand{\vect}[1]{\bm{#1}}
\newcommand{\mat}[1]{\mathbf{#1}}

\newtheorem{remark}{Remark}

\newcommand*{\R}{\mathbb{R}}

\begin{document}

\title{Gridless 2D Recovery of Lines\\ using the Sliding Frank--Wolfe Algorithm}

\author{Kévin Polisano, Basile Dubois-Bonnaire, Sylvain Meignen
\thanks{The authors are with the Jean Kuntzmann Laboratory, University Grenoble Alpes and CNRS 5225, Grenoble 38401, France  (emails: kevin.polisano@univ-grenoble-alpes.fr, basile.dubois-bonnaire@univ-grenoble-alpes.fr, sylvain.meignen@univ-grenoble-alpes.fr).}}

\maketitle
\begin{abstract}
    We present a new approach leveraging the Sliding Frank--Wolfe algorithm to address the challenge of line recovery in degraded images. Building upon advances in conditional gradient methods for sparse inverse problems with differentiable measurement models, we propose two distinct models tailored for line detection tasks within the realm of blurred line deconvolution and ridge detection of linear chirps in spectrogram images.
\end{abstract}
\begin{IEEEkeywords}
superresolution, line detection, convex optimization, Frank--Wolfe algorithm, deconvolution, linear chirps
\end{IEEEkeywords}

\section{Introduction}

\IEEEPARstart{A}{t} the core of numerous imaging applications, the signal of interest can be represented as  a  linear  combination of translated versions of a template \cite{boyd2017alternating}, such as a 
\textit{Point Spread Function} (PSF), for example \cite{denoyelle2019sliding}. The challenge here is to estimate the parameters
encoding the locations and intensities of these templates from measurements corrupted by noise. For instance, in single-molecule
fluorescence  microscopy, the acquisition signal is further constrained by the
aperture of the microscope, resulting in diffraction \cite{pham2021optical}. Therefore, super-resolution techniques aim to
estimate these parameters with greater precision than the Rayleigh limit \cite{candes2014towards}. 

An elegant formulation to perform this task,
resulting in convex optimization problems, is based on the concept of \emph{atomic norm} 
\cite{bhaskar2013atomic,condat2020atomic}. Within this general framework of
atomic norm minimization, the sought-after image $\mat{x}$ is supposed
to be a sparse positive combination of elements from an infinite dictionary
$\mathcal{A}$, indexed by continuously varying parameters. Consequently, the regularization function can be selected as the atomic norm $\|\mat{x}\|_\mathcal{A}$ of the image $\mat{x}$, which
can be viewed as the $\ell_1$ norm of the coefficients of the combination, when the image is
expressed in terms of the sum of unit-norm elements of $\mathcal{A}$, called
\emph{atoms}. In \cite{polisano2019Convex, polisano2016Convex}, the atoms
were \emph{lines}, expressed in the Fourier domain, characterized by their rows and columns, thereby reducing the problem to finding a decomposition in a dictionary of 1-D complex exponential samples, whose frequency and phase encoded the line parameters. The atomic norms of these different slices (extracted rows and columns of the line atoms) were computed via semidefinite programming \cite{vandenberghe1996semidefinite}. This formulation yielded a constrained convex optimization problem, solved using a primal-dual splitting algorithm \cite{condat2013}. Subsequently, by employing a Prony-like method \cite{rahman1987total} on the output of the algorithm, one could extract the parameters of the lines. 

Although this two-step approach is very efficient at estimating lines, 
the convergence of the algorithm used in the first step is particularly slow, 
and the second step is prone to instability when estimating parameters too
closely spaced in the presence of residual noise. It is also worth noting that atomic norm techniques primarily address denoising problems and is not specifically designed to recover the parameters of the underlying signal, as emphasized in \cite{boyd2017alternating}. The authors in this paper argue that this can be circumvented by resorting to the original formulation of the estimation problem as an optimization problem over the space of measures. The parameters are then encoded in the support and amplitude of these measures. Exploiting the structure of the parameter space enabled them to solve a general sparse inverse problem by using and extending the \textit{Conditional Gradient Method} (CGM), also known as \textit{Frank--Wolfe} (FW) algorithm. The resulting \textit{Alternating Conditional Gradient Method} (ACGM) adds a non-convex search step by making local movements in parameter space \cite{boyd2017alternating}. 

This paper explores this framework for achieving super-resolution of lines in images, using the \textit{Sliding Frank-Wolfe} (SFW) algorithm  \cite{denoyelle2019sliding} to solve the Beurling LASSO (BLASSO) problem briefly exposed in Section \ref{sec:blasso}. The first model, detailed in Section \ref{sec:GL}, consists in reformulating the problem of line superresolution introduced in \cite{polisano2019Convex, polisano2016Convex} in this new framework. The \textit{pattern} in this model is a \textit{Gaussian Line} (GL); that is a kernel formed by the convolution of a line distribution with a PSF. The second model, outlined in Section \ref{sec:CL}, will explore a slightly different kernel shape, called \textit{Chirp Line} (CL), specifically designed for detecting linear chirps in spectrogram images. Section \ref{sec:results} discusses the numerical results obtained and compare them with those achieved in \cite{polisano2019Convex, polisano2016Convex}. Finally, a concluding Section \ref{sec:conclusion} addresses the envisioned perspectives.

\section{BLASSO and Frank--Wolfe Algorithms}\label{sec:blasso}

\subsection{Beurling LASSO}

The sparse spikes problem involves reconstructing $\delta$-spikes located in a domain $\mathcal{X} \subset \mathbb{R}^d$, $d\in \mathbb{N}^{\ast}$, from an acquisition $y$, using a prior assumption about the sparsity of the source. The challenge lies in estimating the number $K$ of sources, their positions $\mat{x}=(\vect{x}_k)_{k=1}^K$ and their amplitudes $\vect{\alpha}=(\alpha_k)_{k=1}^K$. The objective is to reconstruct the measure $m_{\vect{\alpha}, \mat{x}}=\sum_{k=1}^K \alpha_k \delta_{\vect{x}_k}$ lying in the Radon measures space $\mathcal{M}(\mathcal{X})$ from a small number of observations $y$ living in a separable Hilbert space $\mathcal{H}$, related to $m_{\vect{\alpha}, \mat{x}}$ through an operator $\Phi: \mathcal{M}(\mathcal{X})\to \mathcal{H}$ modelling the acquisition process as follows:
\begin{equation}\label{eq:operator}
  \forall m\in \mathcal{M}(\mathcal{X}), \quad \Phi m = \int_{\mathcal{X}} \varphi(\vect{x}) \mathrm{d}m(\vect{x}),
\end{equation}
where $\varphi$ denotes the \textit{kernel}, which must be continuous and bounded to ensure that the Bochner integral above is well-defined\footnote{Remark that, for $\mathcal{H}=\mathrm{L}^2(\mathbb{R}^d)$ we have $\varphi(\vect{x})\in \mathcal{H}$ thus for a fixed $\vect{x}\in \mathbb{R}^d$ it is important to note that $\varphi(\vect{x})$ is a function in $\mathrm{L}^2(\mathbb{R}^d)$.}.
For a discrete measure $m=m_{\vect{\alpha}, \mat{x}}$, observations are modeled as:
\begin{equation}\label{eq:atomsum}
    y = \Phi m_{\vect{a}, \mathbf{x}} + w = \sum_{k=1}^K \alpha_k \varphi(\vect{x}_k) + w \;,
\end{equation}
where $w$ represents additive noise. 

The term ``superresolution" indicates that the positions $(\vect{x}_k)_{k=1}^K$ are not constrained to be on a grid but estimated continuously over $\mathcal{X}$. The sparsity prior is enforced by a norm on $\mathcal{M}(\mathcal{X})$ called the total variation (TV) norm, leading to the following variational problem known as BLASSO:

\begin{equation}\label{eq:blasso}\tag{$\mathcal{P}_{\lambda}(y)$}
  \min_{m\in \mathcal{M}(\mathcal{X})} T_\lambda(m)=\| y - \Phi m\|_{\mathcal{H}}^2 + \lambda |m|(\mathcal{X}),
\end{equation}
where $|m|(\mathcal{X})$ is the TV norm of $m$ and $\lambda>0$ is a parameter that needs to be tuned with respect to the noise level $\|w\|$. For a discrete measure $m_{\vect{\alpha}, \mat{x}}$, its TV norm reduces to $\|\vect{\alpha}\|_1$, where the $\ell_1$ norm is known to induce sparsity. 
\begin{remark}
    The relationship between the TV norm and the atomic norm is the following: by considering the dictionary $\mathcal{A}=\{\varphi(\vect{x}):\vect{x}\in \mathcal{X}\}\cup \{\vect{0}\}$, we have the equality 
\begin{equation}
    \|z\|_{\mathcal{A}}= \inf \left\{ |m|(\mathcal{X}) : z=\int_{\mathcal{X}} \varphi(\vect{x})\diff m(\vect{x}),\; m\geq 0\right\}.
\end{equation}
\end{remark}

An optimal solution of \eqref{eq:blasso} is achieved for a measure $m_{\lambda}$ such that the dual certificate\footnote{Applying the Fermat’s rule leads to $m_{\lambda}:=m_{\vect{\alpha},\mat{x}}=\sum_{k=1}^K \alpha_k \delta_{\vect{x}_k}$ solution of $\mathcal{P}_{\lambda}(y)$ if and only if $\eta_{\lambda}=\Phi^{\ast}p_{\lambda}\in \partial |m_{\vect{\alpha},\mat{x}}|$. Then using the expression of the subdifferential of the total variation norm for a discrete measure $m_{\vect{\alpha},\mat{x}}$ gives the expected result $p_{\lambda}=y-\Phi m_{\vect{\alpha},\mat{x}}$ \cite{denoyelle2019sliding}.} $\eta_{\lambda}=\frac{1}{\lambda}\Phi^{\ast}(y-\Phi m_{\lambda})$ satisfies $\|\eta_{\lambda}\|_{\infty, \mathcal{X}}\leq 1$ \cite{denoyelle2019sliding}.

\subsection{The Frank--Wolfe Algorithm}

The Frank--Wolfe (FW) algorithm (see Algorithm \ref{alg:fw}) offers a method for minimizing a convex and differentiable function $f$ over a weakly compact convex set $C$ of a Banach space. At each iteration, the algorithm minimizes  a linear approximation of $f$ over $C$ and constructs the next iterate as a convex combination of the obtained point and the current iterate. The algorithm thus only requires the directional derivatives of $f$ and does not rely on any Hilbertian structure. Even though the functional $T_\lambda$ is not differentiable, it is possible to reframe the problem \eqref{eq:blasso} within this framework by employing an epigraphical lift involving $f:= \tilde T_{\lambda}(m,t)=\| y - \Phi m\|_{\mathcal{H}}^2 + \lambda t$ defined on $C=\{ (m,t): |m|(\mathcal{X}) \leq t \leq r\}$ with $r=\|y\|^2/(2\lambda)$, which is differentiable on the Banach space $\mathcal{M}(\mathcal{X})\times \mathbb{R}$ with differential:
\begin{equation}\label{eq:Tdiff}
  \mathrm{d}\tilde T_{\lambda}(m,t): (m',t')\mapsto \int_{\mathcal{X}} \Phi^{\ast}(\Phi m-y)\diff m' + \lambda t'.
\end{equation}
The FW algorithm is useful only if in step 2 of Algorithm \ref{alg:fw} one is able to minimize the linear form $s\mapsto \mathrm{d}\tilde T_{\lambda}(m,t)[s]$ on $C$ \cite{denoyelle2019sliding}. It reaches its minimum at least at  an extreme point of $C$, i.e $s=(0,0)$ or $s=r\cdot(\pm \delta_{\vect{x}},1)$ for $\vect{x}\in \mathcal{X}$. Thus, at iteration $j$, for a current measure $m^j=\sum_{k=1}^{j} \alpha_k^j\delta_{\vect{x}_k^j}$ one has to find $\vect{x}_{\ast}^j\in \argmax_{\vect{x}\in \mathcal{X}} |\eta^j(\vect{x})|$ where $\eta^j = \frac{1}{\lambda}\Phi^{\ast}(y-\Phi m^j)$. This greedy step 2 of Algorithm \ref{alg:fw}, first initialized with a grid search, aims at creating a new spike at each iteration. 
When FW algorithm stops, the quantity $\eta^j$ is the dual certificate $\eta_{\lambda}$ for $\mathcal{P}_{\lambda}(y)$, satisfying $\eta^j(\vect{x}_k^j)=\mathrm{sign}(\alpha_k^j)$. 
Another noteworthy aspect of the FW algorithm is its flexibility in the final update step, where the point $m^{j+1}$ can be replaced by any point $\hat{m} \in C$ satisfying $f(\hat{m}) \leq f(m^{j+1})$, while preserving all convergence properties. This degree of freedom has spurred interesting enhancements of FW algorithm \cite{boyd2017alternating,denoyelle2019sliding}. In these works, the final update step is modified to involve a non-convex optimization problem, updating both the amplitudes and positions of the spikes to further minimize the objective function. The \textit{Sliding Frank--Wolfe} algorithm (SFW) \cite{denoyelle2019sliding} improved upon the \textit{Alternating Conditional Gradient Method} (ACGD) \cite{boyd2017alternating} by simultaneously considering both the amplitudes and positions of the reconstructed spikes, aiming to achieve convergence within a finite number of iterations and enhance the robustness of the estimations. We denote by $K_{\mathrm{max}}$ the maximum number of iterations performed by the algorithm.

\begin{algorithm}[ht]
  \caption{Frank--Wolfe Algorithm}
  \label{alg:fw}
  \begin{algorithmic}[1]
  \FOR {$j=1$ \TO $K_{\mathrm{max}}$} 
  \STATE Minimize: $\displaystyle s^j\in \argmin_{s\in C} f(m^j)+\mathrm{d}f(m^j)(s-m^j)$
  \IF{$\mathrm{d}f(m^j)(s^j-m^j)=0$} 
  \STATE output $m^{\star}\leftarrow m^j$ is optimal.
  \ELSE
  \STATE $\gamma^j\leftarrow \frac{2}{j+2}$ or $\displaystyle \gamma^j\leftarrow \argmin_{\gamma \in [0,1]} f(m^j+\gamma (s^j-m^j))$
  \STATE Update: $m^{j+1}\leftarrow m^j + \gamma^j (s^j-m^j)$
  \ENDIF 
  \ENDFOR
  \end{algorithmic}
  \end{algorithm}

\section{Gaussian Lines Model}\label{sec:GL}

\subsection{Convolution of a Line with a Point Spread Function}

A perfect line is characterized by an angle $\theta\in (-\pi/2,\pi/2]$ relative to
the vertical axis, amplitude $\alpha>0$ and offset $\eta\in\mathbb{R}$ from
the origin along the horizontal axis (see \cite[Figure 2]{polisano2019Convex}). This ideal line can be defined as the tempered distribution
$\delta_{\mathscr{L}(\eta,\theta)}$, which maps a function $\psi$ in the Schwartz class
$\mathcal{S}( \mathbb{R}^2)$ to its integral along the geometric line $
\mathscr{L}(\eta,\theta)=\{(u_1,u_2)\in\mathbb{R}^2: (u_1-\eta)\cos \theta +u_2\sin
\theta=0\}$. 

The convolution of the distribution $\delta_{\mathscr{L}(\eta,\theta)}$ with a point spread function (PSF) $\phi\in \mathcal{S}(\mathbb{R}^2)$ is given
by: 
\begin{equation}\label{eq:delta_L2}
    (\delta_{\mathscr{L}(\eta,\theta)}\ast \phi)(\vect{u})=\langle \delta_{\mathscr{L}(\eta,\theta)},\phi(\vect{u}-\cdot)\rangle
=\int_{\mathscr{L}(\eta,\theta)}\phi(\vect{u}-\vect{v})\diff \vect{v}.
\end{equation}
 For a separable PSF $\phi(u_1,u_2)=\phi_1(u_1)\phi_2(u_2)$
we have 
\begin{equation}\label{eq:conv}
  (\delta_{\mathscr{L}(\eta,\theta)}\ast
\phi)(u_1,u_2) = \phi_{\theta}((u_1-\eta)\cos \theta + u_2\sin \theta),
\end{equation}
where $
  \phi_{\theta}=\frac{1}{\cos \theta \sin \theta}\phi_1\Big(\frac{\cdot}{\cos \theta}\Big)\ast \phi_2\Big(\frac{\cdot}{\sin \theta}\Big)$, moreover if \begin{equation}\label{eq:gauss}
  \phi_i(u)=\frac{1}{\sqrt{2\pi
\sigma_i^2}}\exp\left(-\frac{u^2}{2\sigma_i^2}\right), \quad i\in\{1,2\},
\end{equation}
then with $\sigma_{\theta}^2=\sigma_1^2\cos^2 \theta
+\sigma_2^2\sin^2 \theta$ we obtain
\begin{equation}\label{eq:phit}
    \phi_{\theta}(u)=\frac{1}{\sqrt{2\pi \sigma_{\theta}^2}}\exp\left(-\frac{u^2}{2\sigma_{\theta}^2}\right).
\end{equation}

\begin{figure}
    \centering
     \subfloat[]{
    \includegraphics[width=0.23\textwidth, height=0.23\textwidth]{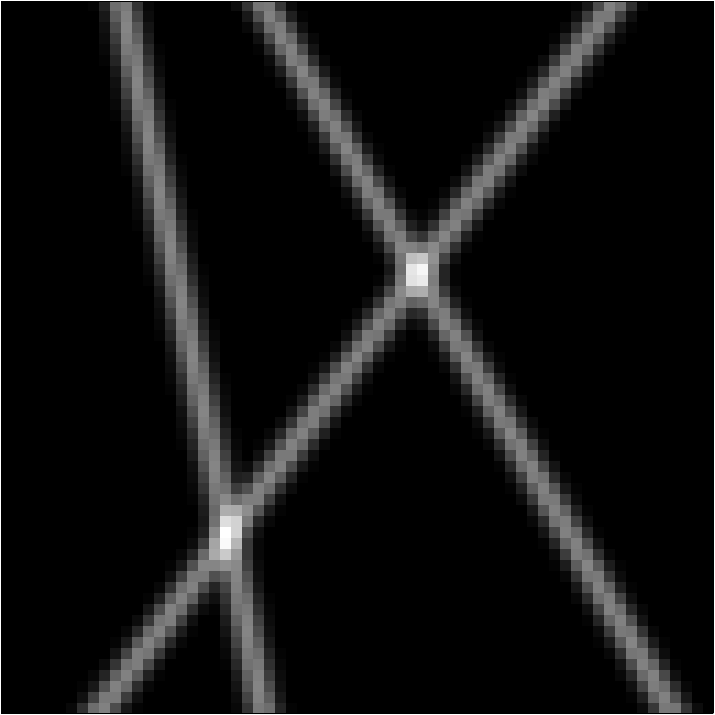}\label{figa:Radon}
    }
     \subfloat[]{
    \includegraphics[width=0.23\textwidth, height=0.23\textwidth,trim={30 30 50 20},clip]{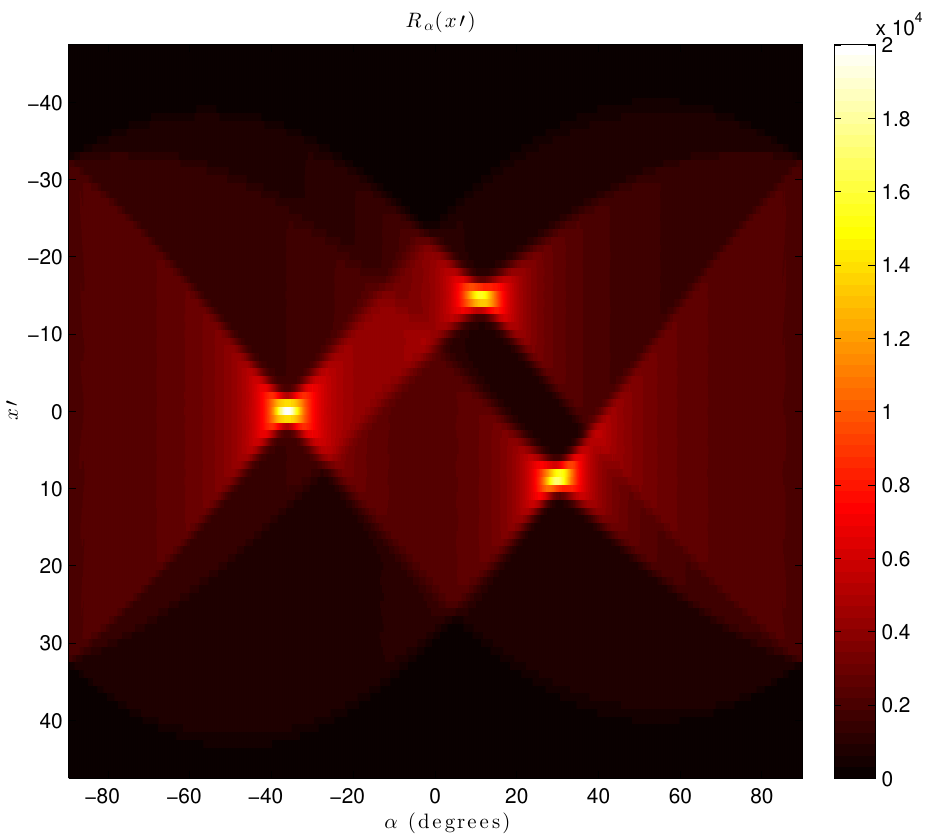}\label{figb:Radon}}
    \caption{(a) Three Gaussian Lines with parameters of Experiment 1 (see Section \ref{sec:GLest}) in the noiseless case. (b) Its Radon transform in the plane $(\theta,\eta)$.}
    \label{fig:Radon}
\end{figure}

\subsection{Gaussian Line Kernel}
For a given $\vect{x}=(\eta,\theta)\in \mathcal{X}$ where $\mathcal{X}=\mathbb{R}\times \left(-\frac{\pi}{2},\frac{\pi}{2}\right]\subset \mathbb{R}^2$, we consider a separable Gaussian PSF $\phi$ defined by equation \eqref{eq:gauss}, and we define the \textit{Gaussian Line} kernel $\varphi_{\mathrm{GL}}:\mathcal{X}\to \mathcal{H}$ as $\varphi_{\mathrm{GL}}(\vect{x})=\delta_{\mathscr{L}(\vect{x})}\ast \phi$. In other words, using equations \eqref{eq:conv}-\eqref{eq:phit} we obtain the explicit formula:
\begin{equation}\label{eq:kernel1}
  \varphi_{\mathrm{GL}}(\eta,\theta):= (u_1,u_2)\mapsto \phi_{\theta}\left((u_1-\eta)\cos(\theta)+u_2\sin(\theta)\right).
\end{equation}
The kernel is sampled with unit step onto a grid $\llbracket -M,M\rrbracket^2$, resulting in a discrete image of size $N\times N$ with $N=2M+1$. A sum of three Gaussian lines is plotted in Figure \ref{figa:Radon}.  We consider $\mathcal{H}=\mathbb{R}^{N^2}$ by vectorizing  this image into a vector.
\begin{remark}
  For $\mathcal{H}=\mathrm{L}^2(\mathbb{R}^2)$ the kernel \eqref{eq:kernel1} does not satisfy the admissibility kernel assumptions stated in \cite[Definition 4]{denoyelle2019sliding}, since $\|f\|_{\mathcal{H}}^2=\int_{\mathbb{R}^2} f^2$ and for a fixed $\boldsymbol{x}=(\eta,\theta)$, the integral of $\varphi_{\mathrm{GL}}(\vect{x})$ along the line $\mathscr{L}(\eta,\theta)$ is infinite\footnote{One could apply a window with a smooth decreasing towards zero in order to overcome this issue.}. However, in the discrete setting with $\mathcal{H}=\mathbb{R}^{N^2}$, these assumptions are satisfied since the kernel vanishes when $\eta\to \infty$. 
\end{remark}

\section{Linear chirps as Lines in the Spectrogram}\label{sec:CL}

\subsection{Spectrogram of the superimposition of linear chirps}

To capture the frequency variations of a signal over time, the \emph{Short-Time Fourier Transform} (STFT) is commonly used. For a signal $f$ and a real window $h$ both in $L^2(\R)$, the STFT is defined as follows: 
\begin{eqnarray}
\label{def:STFT}
V_f^h(t, \omega) = \int_\R f(s) h(s-t) \eu^{-i 2\pi \omega (s-t)} \diff s.
\end{eqnarray}
The \emph{spectrogram} is then defined as the squared modulus of the STFT. 
In the following, we will use a Gaussian window $h_\sigma(t) = \eu^{-\pi \frac{t^2}{\sigma^2}}$. We consider a discretized version of the STFT in both time and frequency, 
namely one defines ${\bm V}_f^h[n,k] \approx V_f^h(\frac{n}{N},k)$, the indices $n$ and $k$ both varying in $\llbracket 0,N-1\rrbracket$.

Now, considering a \textit{linear chirp} $f_{\eta,\beta}(t)=\eu^{2i\pi \left(\eta t + \frac{\beta}{2}t^2\right)}$, one can show that the corresponding spectrogram has the following explicit formula \cite{meignen2022one}:
\begin{equation}
\label{def:spec}
  |V_{f_{\eta,\beta}}^h(t, \omega)|^2 = \frac{\sigma^2}{\left(1+\sigma^4\beta^2\right)^{\frac{1}{2}}}\eu^{-\frac{2\pi \sigma^2 (\omega-(\eta+\beta t))^2}{1+\sigma^4\beta^2}}\;.
\end{equation}
This expression reveals a \textit{ridge} along the line $\omega-(\eta+\beta t)=0$, with a Gaussian profile whose variance depends on the slope $\beta$ of the line.
For a superimposition of $K$ linear chirps $f=\sum_{k=1}^K f_{\eta_k, \beta_k}$, the corresponding spectrogram is the sum of the spectrograms of each $f_{\eta_k,\beta_k}$ plus some interference cross-terms leading to oscillating pattern sometimes referred to as time-frequency \textit{bubbles} \cite{meignen2022one}. If the ridges are sufficiently far apart, these interference are weak, but otherwise they can degrade the spectrogram. Although we neglect these interference in the modeling, we test the line estimation in the presence of such interference in Section \ref{sec:results}.

\begin{figure*}
        \subfloat[]{ \includegraphics[clip, trim=76px 51px 38px 77px, width=0.23\textwidth, height=0.23\textwidth]{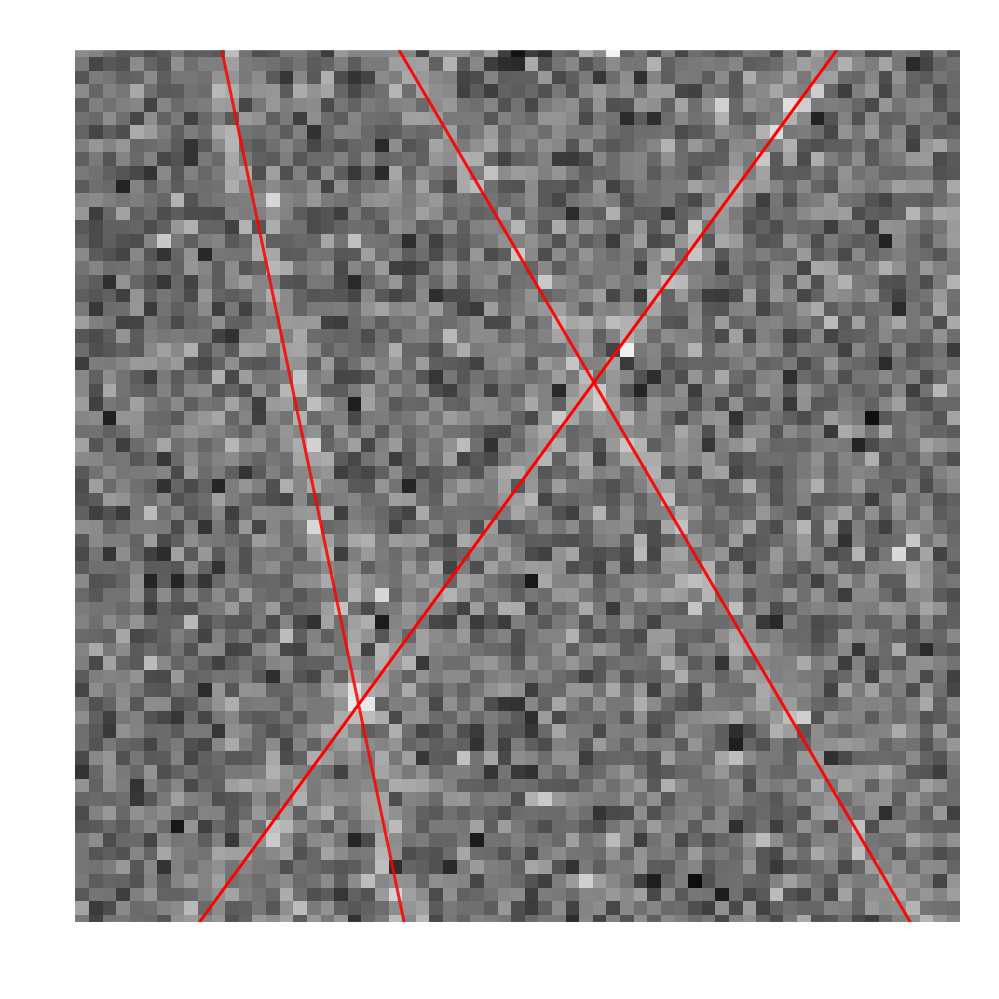}\label{figa:gaussian}
    }
    \subfloat[]{
\includegraphics[clip, trim=76px 51px 38px 77px,width=0.23\textwidth, height=0.23\textwidth]{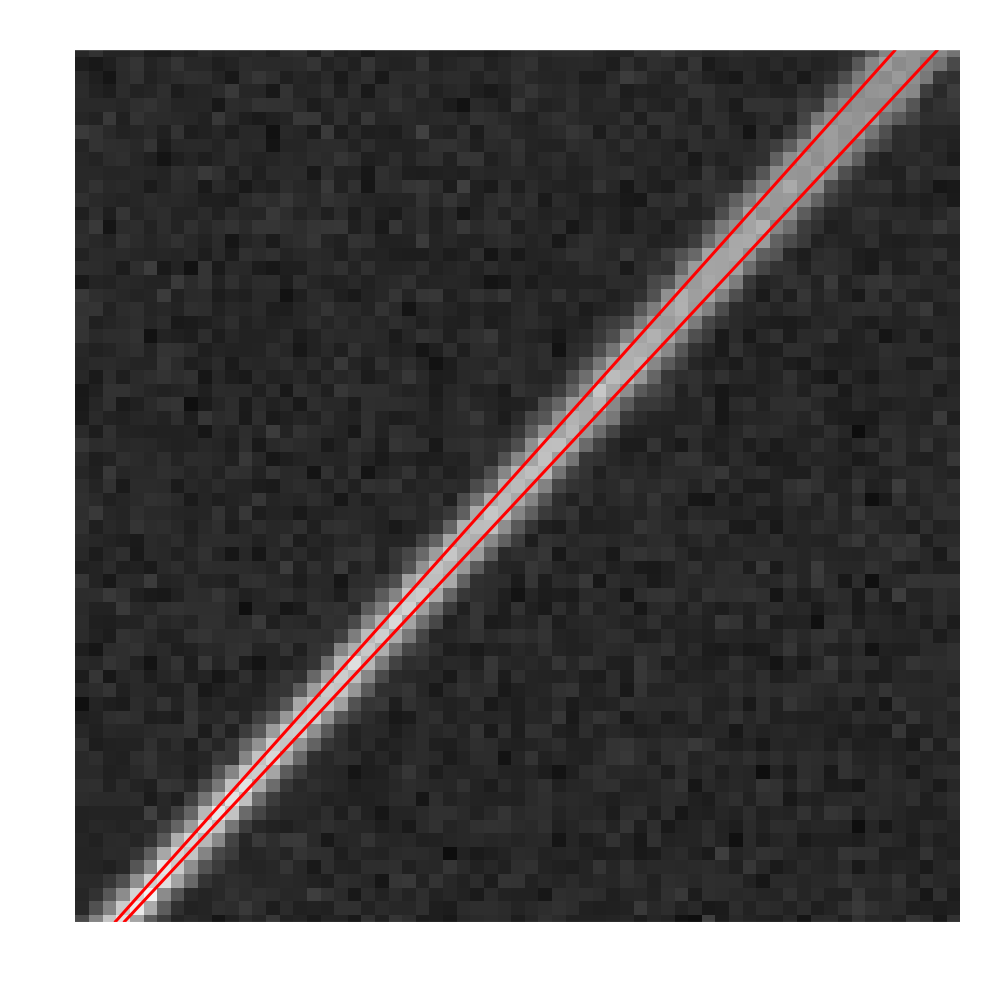}\label{figb:gaussian}
    } 
     \subfloat[]{
\includegraphics[clip, trim=76px 51px 38px 77px,width=0.23\textwidth, height=0.23\textwidth]{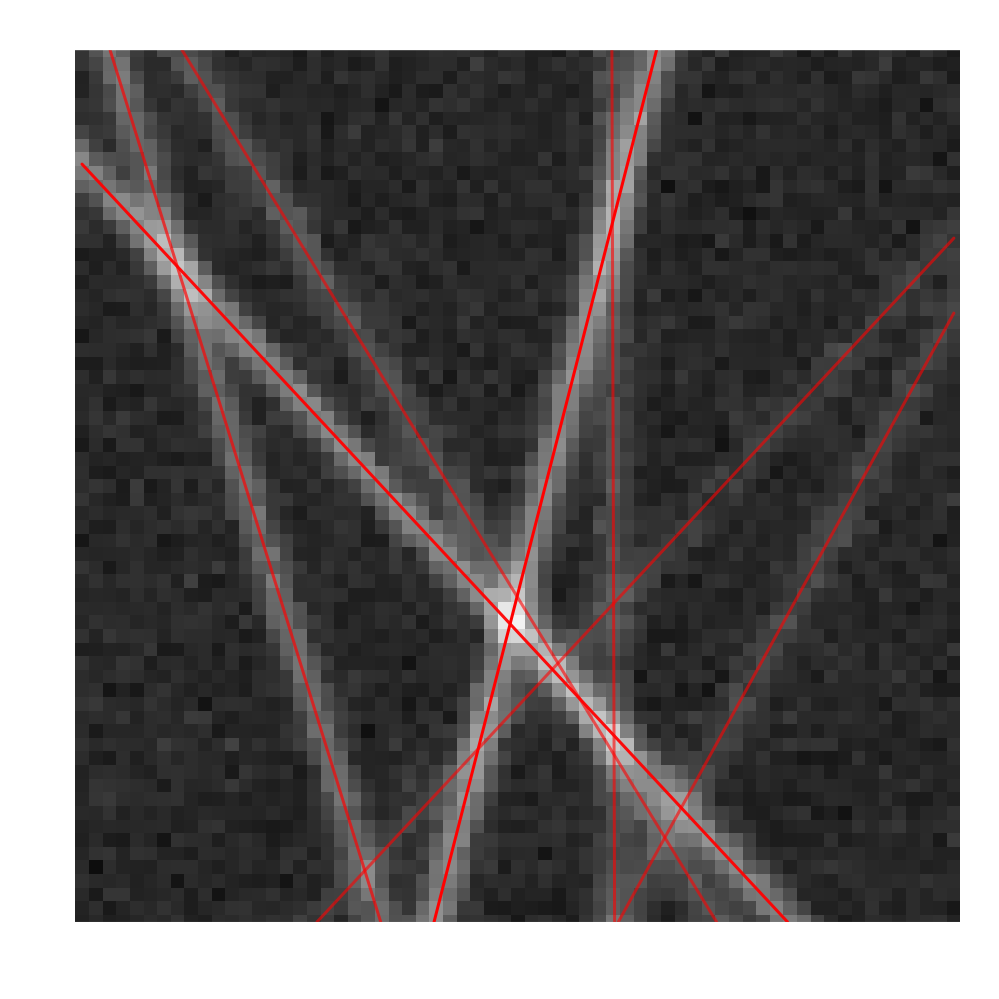}\label{figc:gaussian}}
     \subfloat[]{
\includegraphics[trim=76px 51px 38px 77px,width=0.23\textwidth, height=0.23\textwidth]{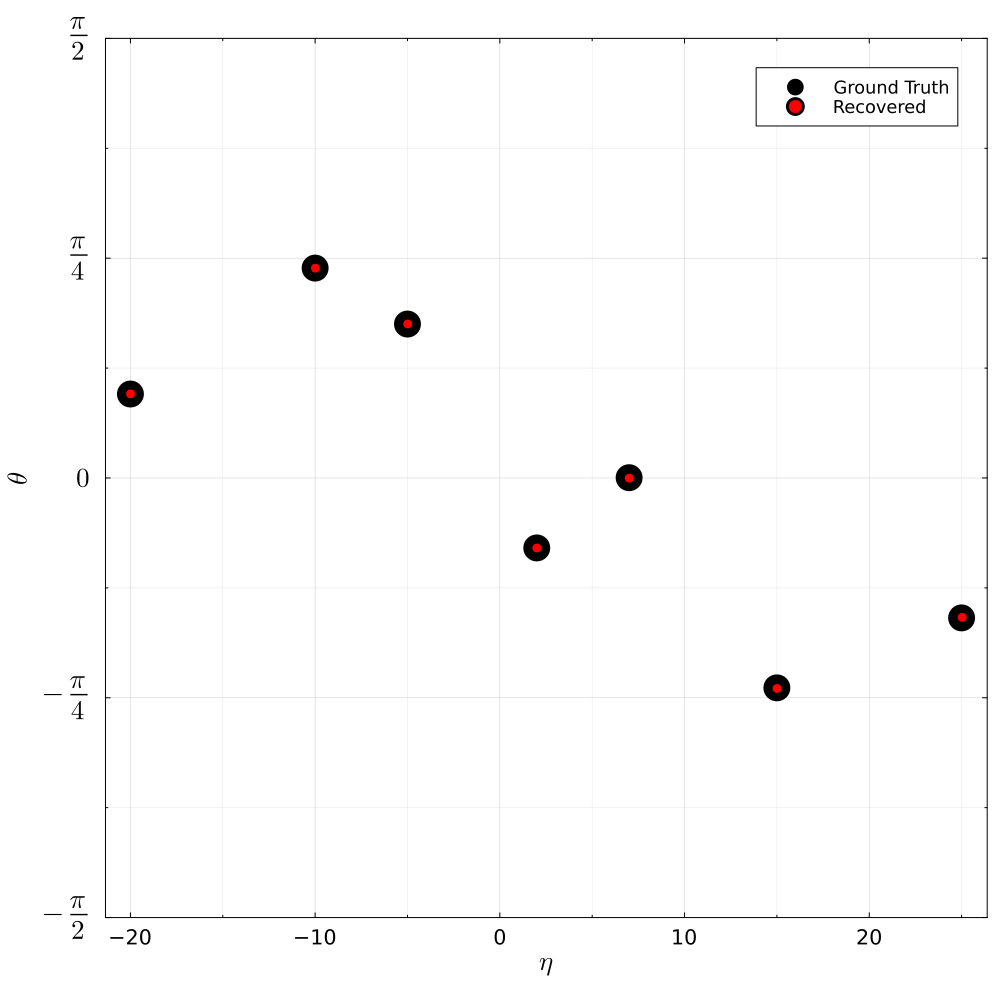}\label{figd:gaussian}}
    \caption{(a) Exp. 1 (very noisy lines), (b) Exp. 2 (very close lines plus noise), (c) Exp. 3 (more lines with different amplitudes plus noise). The estimated lines are depicted in red. (d) Ground truth spikes (in black) and estimated ones (in red) in the parameter space.}
    \label{fig:gaussian}
\end{figure*}

\subsection{Chirp Line Kernel}\label{sec:chirp_kernel}

We parametrize a line with two parameters $\boldsymbol{x}=(\eta,\theta)$,  as previously. 
We consider $\beta = \tan \theta$, and based on \eqref{def:spec}, we consider $\mathcal{H}=\mathbb{R}^{N^2}$ and then define the \textit{chirp line} kernel $\varphi_{\mathrm{CL}}:\mathcal{X}\to \mathcal{H}$ as the discretization of the function:
\begin{equation}\label{eq:opCL}
  \varphi_{\mathrm{CL}}(\eta,\theta):=  (t,\omega)\mapsto \sigma_N(\theta)\,\eu^{-\frac{2\pi \sigma^2 (\omega-(N-1)(\eta+\tan(\theta) t))^2}{1+\sigma^4(N-1)^2\tan^2\theta}},
\end{equation}
where $\sigma_N(\theta)=\sigma^2\left(1+\sigma^4(N-1)^2\tan^2\theta\right)^{-\frac{1}{2}}$ and $[0,1]^2$ uniformly sampled onto a grid $N\times N$ and vectorized\footnote{Remark that the kernel \eqref{eq:opCL} has a slightly different expression compared to the chirp spectrogram \eqref{def:spec}, involving a constant $N$. This adjustment ensures that the frequencies in the spectrogram lie within the range $\llbracket 0, N-1\rrbracket$ when sampling the kernel onto $[0,1]^2$. This way the angles span $\left[-\frac{\pi}{2},\frac{\pi}{2}\right]$, whereas onto $[0,1]\times [0,N-1]$ they would be concentrated near $\pm \frac{\pi}{2}$.}.
\begin{remark}
The kernel satisfies the admissibility kernel assumptions \cite[Definition 4]{denoyelle2019sliding}. Indeed: 
$$\|\varphi_{\mathrm{CL}}(\boldsymbol{x})\|_{2}^2=\sum_{n=1}^{N^2} [\varphi_{\mathrm{CL}}(\boldsymbol{x})]_{n}^2\leq N^2 \sigma^2 \left(1+\sigma^4\tan^2\theta\right)^{-1}\;,$$
so for $\epsilon>0$, there exists a compact set $\mathcal{K}=\mathcal{K}_1\times \mathcal{K}_2\subset \mathcal{X}$, where $\mathcal{K}_2=[-\vartheta_{\epsilon},\vartheta_{\epsilon}]\subset \left(-\frac{\pi}{2},\frac{\pi}{2}\right]$ such that $\sup_{\boldsymbol{x}\in \mathcal{X}\backslash \mathcal{K}}\|\varphi_{\mathrm{CL}}(\boldsymbol{x})\|_{2}\leq \epsilon$ since $\lim_{\theta\to \pm \frac{\pi}{2}}\tan(\theta)=\pm \infty$. The second assumption which requires the function $\psi_{\vect{p}}(\vect{x})=\langle \varphi_{\mathrm{CL}}(\vect{x}), \vect{p}\rangle_{\mathbb{R}^{N^2}}$
to vanish at the limits of the domain (when $\theta\to \pm \pi/2$) is satisfied for all $\vect{p}\in \mathbb{R}^{N^2}$, since by Cauchy-Schwarz inequality one  has $ |\psi_{\vect{p}}(\boldsymbol{x})|\leq \|\varphi_{\mathrm{CL}}(\boldsymbol{x})\|_{2} \|\vect{p}\|_{2}$ and the result follows by taking the supremum.
\end{remark}

\section{Numerical experiments and Discussions}\label{sec:results}

\subsection{Implementation details}

We are dealing with images of size $N\times N$ containing $K$ lines degraded through an operator $\Phi$ defined in \eqref{eq:operator}. In our case, the associated kernel is either $\varphi_{\mathrm{GL}}$ or $\varphi_{\mathrm{CL}}$. In both scenarios, we want to estimate $K$ amplitudes $\vect{\alpha}=(\alpha_k)_{k=1}^K$ and $K$ parameters $\mathbf{x}=(\vect{x}_k)_{k=1}^K$, where $\vect{x}_k=(\eta_k, \theta_k)$, supported by the discrete measure $m_{\vect{\alpha}, \mathbf{x}}=\sum_{k=1}^K \alpha_k \delta_{\vect{x}_k}$, from the observations $\vect{y}\in \mathbb{R}^{N^2}$ obtained from equation \eqref{eq:atomsum}.

\begin{remark}
In both models, we chose an angle parameter $\theta$ instead of the slope of the line. This choice allows us to sample the parameter within a compact set $[\theta_{\mathrm{min}}, \theta_{\mathrm{max}}]\subset \left(-\frac{\pi}{2},\frac{\pi}{2}\right]$. Then, concerning the offset parameter, one is now able to compute $[\eta_{\mathrm{min}}, \eta_{\mathrm{max}}]$ which represent the limits beyond which the lines are no longer visible within the image. For the second model, we employ a slightly different parametrization compared with the first one (see \cite[Figure 2]{polisano2019Convex}), where the angle $\theta$ is taken with respect to the $x$-axis\footnote{This choice is motivated by the observation that the chirp lines in the spectrogram images cannot be ``vertical", the slope $\beta$ being bounded.}, and $\eta$ represents the offset along the $y$-axis aligned with the left border of the image rather than passing through its center in the first model. 
\end{remark}

\textbf{Greedy step enhancement}. The resulting parameter space $\mathcal{X}=[\theta_{\mathrm{min}}, \theta_{\mathrm{max}}]\times [\eta_{\mathrm{min}}, \eta_{\mathrm{max}}]$ was initially uniformly sampled onto a grid of size $P\times P$. This grid was used in the greedy grid search (step 2 of Algorithm \ref{alg:fw}) to find the maximum of the correlations $\eta^j(\vect{x}_p)$ among the $P^2$ sample points $\vect{x}_p$ lying on the coarse grid. We improved this step specifically for line detection purposes. We propose to replace this costly step by the search of local maxima within the Radon transform of $\vect{y}$ (see Figure \ref{figb:Radon}). Indeed, the projection of lines produces a peak in the vicinity of the angle and offset to be recovered. This initial coarse estimation of candidate parameters speeds up the greedy procedure. Subsequently, the non-convex step refines this estimation to achieve super-resolution precision.

The accuracy of line estimation is evaluated by computing the average absolute differences $\overline{\Delta_{\theta}}$, $\overline{\Delta_{\eta}}$ and $\overline{\Delta_{\alpha}}$ between the estimated and true parameters for each line (see Table \ref{table_errors}).


The code implementation in Julia and the notebooks to reproduce the following experiments are available on GitHub: \href{https://github.com/bdbonnaire/sfw4blasso}{https://github.com/bdbonnaire/sfw4blasso}.

\subsection{Gaussian Lines Estimation}\label{sec:GLest}

In this setting, images have a dimension of $N=65$ and contained blurred lines (with $\sigma_1=\sigma_2=1$) corrupted by additional white noise $\vect{w}\sim \mathcal{N}(0,\sigma^2)$. We replicated three experiments from \cite{polisano2019Convex} for benchmarking purposes:
\begin{itemize}
    \item \textbf{Exp. 1} \textit{(Very noisy lines)}: $K=3$ lines with  amplitudes equal to 1, $\vect{x}_1=(0,-\pi/5)$, $\vect{x}_2=(-15,\pi/16)$, $\vect{x}_3=(10,\pi/6)$ and $(\sigma,\lambda)=(0.31,10)$.
    \item \textbf{Exp. 2} \textit{(Very close lines)}: $K=2$ lines with  amplitudes equal to 1, $\vect{x}_1=(-1,-0.73)$, $\vect{x}_2=(1,-0.75)$ and  $(\sigma,\lambda)=(0.031,0.5)$.
    \item \textbf{Exp. 3} \textit{(More lines with different amplitudes)} $K=7$ lines: $\vect{x}_1=(15, -0.75)$, $\vect{x}_2=(25, -0.5)$, $\vect{x}_3=(2, -0.25)$, $\vect{x}_4=(7, 0.001)$, $\vect{x}_5=(-20, 0.3)$, $\vect{x}_6=(-5, 0.55)$, $\vect{x}_7=(-10, 0.75)$ and corresponding amplitudes $\vect{\alpha}=(0.23, 0.31, 1.0, 0.39, 0.7, 0.47, 0.94)$, with $(\sigma,\lambda)=(0.031,1)$.
\end{itemize}

\begin{figure*}[t]
    \centering
    \subfloat[]{
    \includegraphics[width=0.23\textwidth, height=0.23\textwidth]{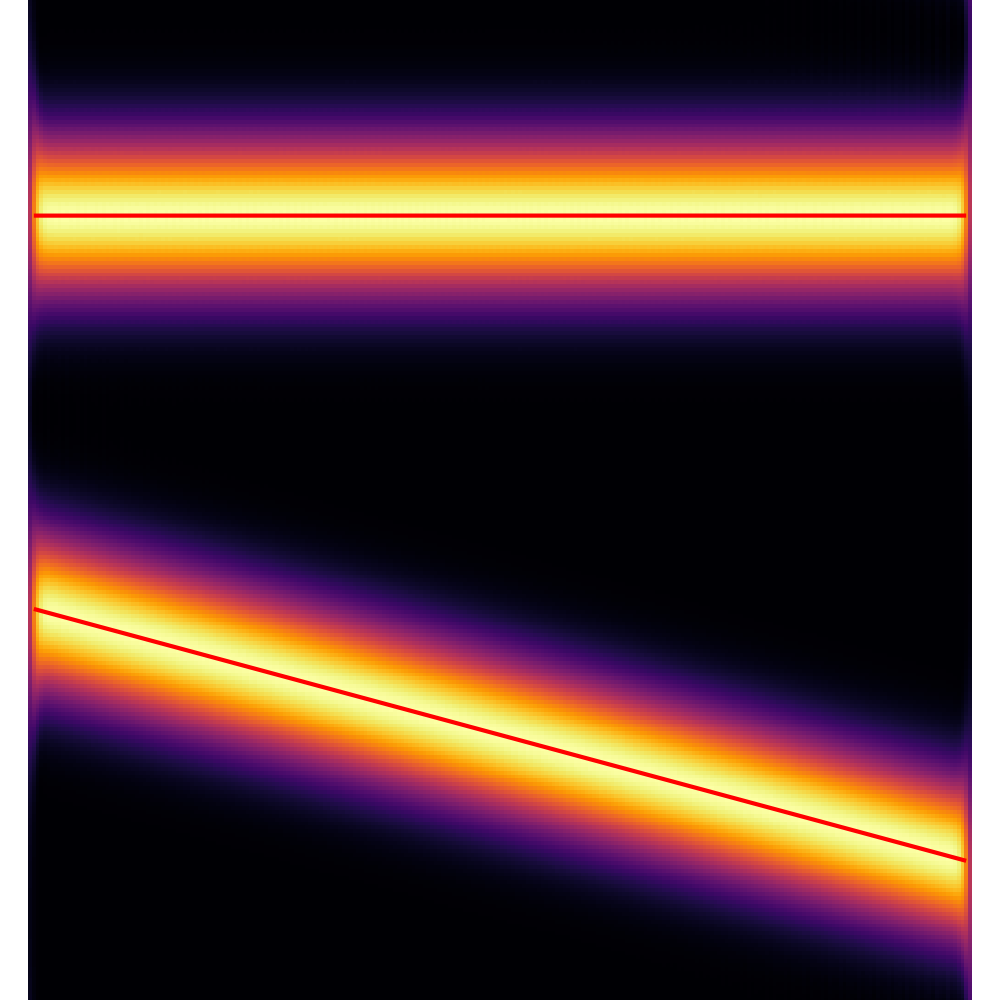}\label{figa:chirps}
    }
     \subfloat[]{
    \includegraphics[width=0.23\textwidth, height=0.23\textwidth]{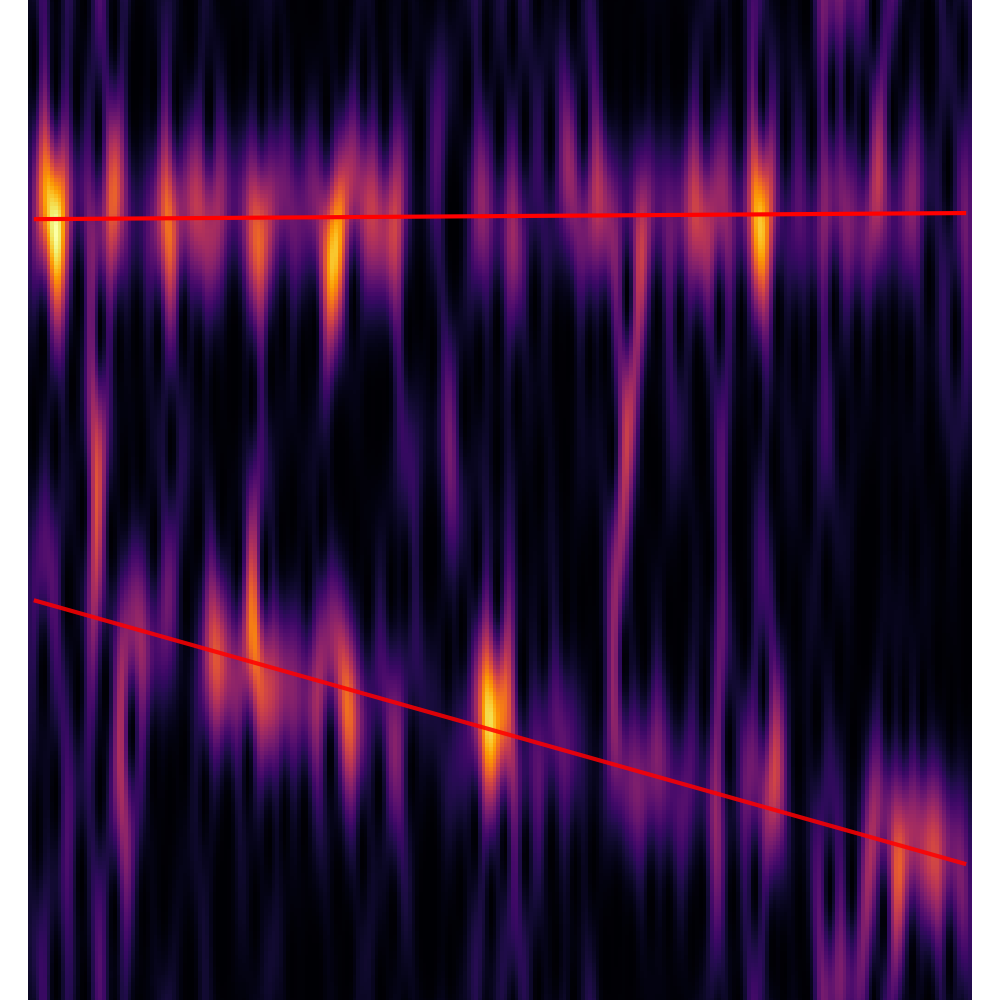}\label{figb:chirps}}
     \subfloat[]{
\includegraphics[width=0.23\textwidth, height=0.23\textwidth]{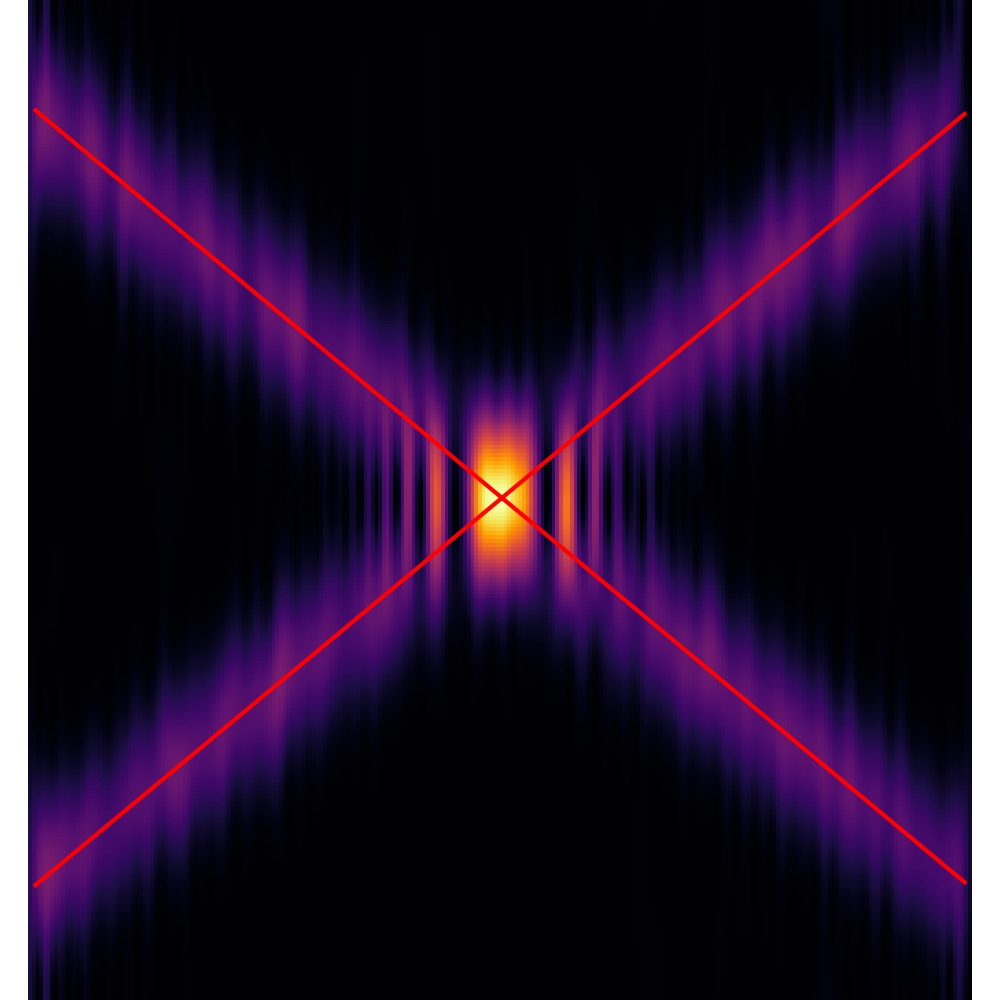}\label{figc:chirps}
    }
     \subfloat[]{
    \includegraphics[width=0.23\textwidth, height=0.23\textwidth]{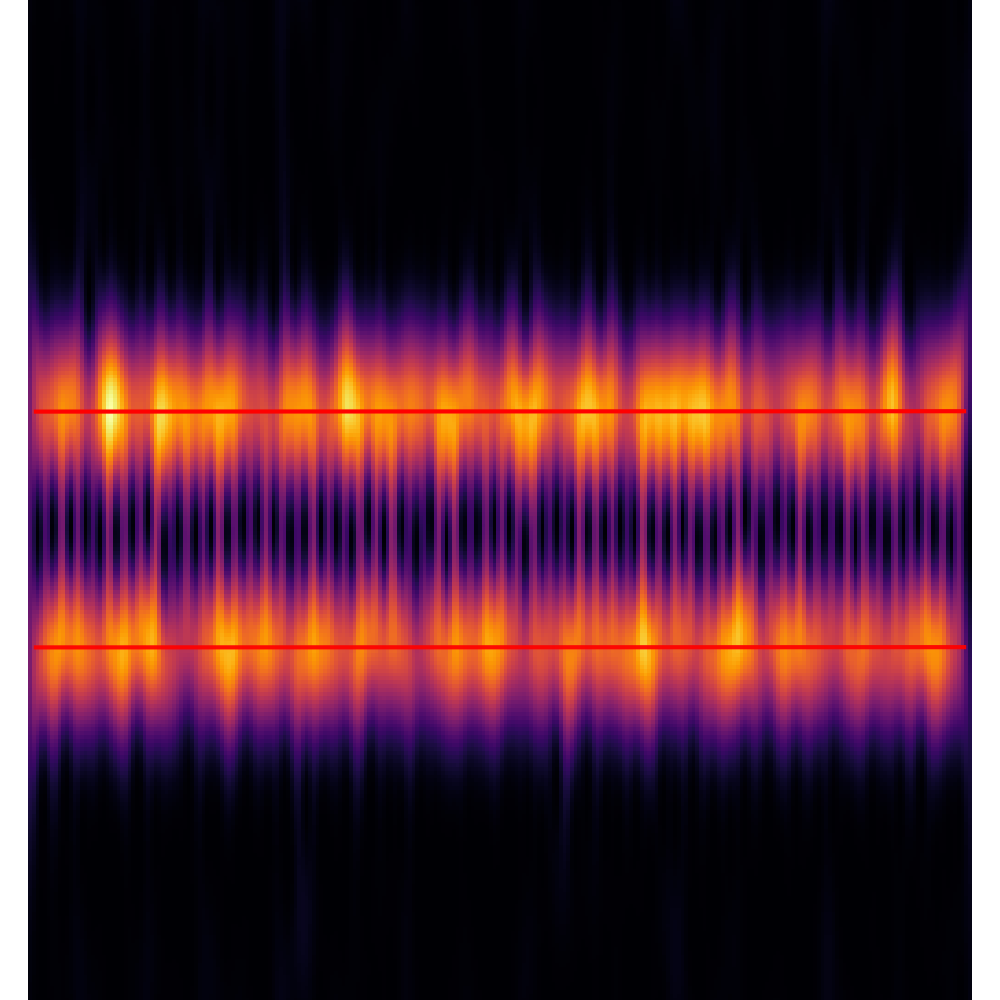}\label{figd:chirps}}
    \caption{(a) Exp. 4 in the noiseless case, (b) Exp. 4 with no interference and high amount of noise ($\chi_2$ distributed), (c) Exp. 5 for crossing lines with interference and moderate noise, (d) Exp. 6 for parallel close lined with more interference and moderate noise. The estimated lines are depicted in red.}
    \label{fig:chirps}
\end{figure*}

\begin{table*}[!t]
\renewcommand{\arraystretch}{1.3}
\caption{Errors on estimated parameters}
\label{table_errors}
\centering
\begin{tabular}{|c|c|c|c|c|c|c|c|}
\cline{2-8}
\multicolumn{1}{c|}{ } & \bfseries Error &\bfseries Exp. 1& \bfseries Exp. 2 &
\bfseries Exp. 3 & \bfseries Exp. 4 & \bfseries Exp. 5 & \bfseries Exp. 6\\
\hline
\multirow{2}{*}[-0.8em]{\cite{polisano2019Convex}}   & $\overline{\Delta_{\theta}}$&\small $2\cdot 10^{-2}$&\small $4\cdot 10^{-3}$&\small $6\cdot 10^{-3}$ & \cellcolor{gray!50} &\cellcolor{gray!50} &\cellcolor{gray!50}\\
\cline{2-8}
 & $\overline{\Delta_{\eta}}$&\small $4\cdot 10^{-2}$&\small $6\cdot 10^{-1}$
&\small $2\cdot 10^{-1}$&\cellcolor{gray!50} &\cellcolor{gray!50} &\cellcolor{gray!50}\\
\cline{2-8}
 & $\overline{\Delta_{\alpha}}$&\small $1\cdot 10^{-1}$ &\small $4\cdot 10^{-2}$
&\small $2\cdot 10^{-2}$&\cellcolor{gray!50} &\cellcolor{gray!50} &\cellcolor{gray!50}\\
\hline
\multirow{2}{*}[-0.8em]{\textbf{Proposed approach}}   & $\overline{\Delta_{\theta}}$& $\bf 1\cdot 10^{-3}$ &\small $\bf 5\cdot 10^{-4}$ &\small $\bf 2\cdot 10^{-4}$ &\small $6\cdot 10^{-3}$ &\small $2\cdot 10^{-3}$ &\small $6\cdot 10^{-4}$\\
\cline{2-8}
 & $\overline{\Delta_{\eta}}$&\small $\bf 2\cdot 10^{-2}$ &\small $\bf 3\cdot 10^{-2}$
&\small $\bf 2\cdot 10^{-2}$ &\small $4\cdot 10^{-3}$ &\small $6\cdot 10^{-4}$ &\small $5\cdot 10^{-4}$\\
\cline{2-8}
 & $\overline{\Delta_{\alpha}}$&\small $\bf 3\cdot 10^{-2}$
 &\small $\bf 8\cdot 10^{-3}$
&\small $\bf 1\cdot 10^{-2}$
&\small $2\cdot10^{-1}$
&\small $4\cdot10^{-2}$
&\small $1\cdot10^{-2}$\\
\hline
\end{tabular}
\end{table*}

\subsection{Chirp Lines Estimation}

In this setting, images have a dimension of $N=256$. A white noise $\vect{w}\sim \mathcal{N}(0,\sigma^2)$ is added to the 1D signal formed by the superposition of $K=2$ chirps with equal amplitudes $\alpha_1=\alpha_2=1$. Note that the spectrogram of $\vect{w}$ is $\chi_2$ distributed with 2 degrees of freedom \cite{pham2018novel}.  We conducted three experiments:
\begin{itemize}
    \item \textbf{Exp. 4} \textit{(Very noisy lines)}: $\vect{x}_1=(0.78, 0)$, $\vect{x}_2=(0.39,-0.25)$ and $(\sigma,\lambda)=(1,0.01)$. 
    \item \textbf{Exp. 5} \textit{(Crossing lines with interference)}:  $\vect{x}_1=(-1,-0.73)$, $\vect{x}_2=(1,-0.75)$ and $(\sigma,\lambda)=(0.031,0.5)$.
    \item \textbf{Exp. 6} \textit{(Parallel close lines with interference)} $\vect{x}_1=(0.89,-0.66)$, $\vect{x}_2=(0.1,0.66)$ and $(\sigma,\lambda)=(0.2,0.01)$. 
\end{itemize}

\subsection{Discussions}

All experiments from 1 to 6 for line detection are shown in this order in Figure \ref{figa:gaussian}, \ref{figb:gaussian}, \ref{figc:gaussian}-\ref{figd:gaussian} and Figure \ref{figa:chirps}-\ref{figb:chirps}, \ref{figc:chirps}, \ref{figd:chirps}; with corresponding parameter estimation errors detailed in Table \ref{table_errors}. The method we propose consistently outperforms \cite{polisano2019Convex} in terms of accuracy for Gaussian Lines detection across the first three experiments. Similarly, for experiments 4, 5, and 6, the accuracy of parameter estimation of the Chirp Lines is notably high. 

Moreover, this new approach has overcome several limitations of \cite{polisano2019Convex}: the choice of the kernel representing the degradation of the lines is no longer limited to convolution with a separable PSF, allowing us to handle Chirp Lines; we are no longer constrained by angles in $[-\pi/4,\pi/4]$ or compelled to split the image in two parts for capturing vertical and horizontal lines separately in a costly optimization problem; the convergence of the algorithm is significantly faster, allowing us to handle larger images; and finally by working directly in the parameter space, we no longer need to operate in two steps: denoising and deconvolution via the atomic norm, followed by Prony methods to estimate the line parameters, which was more unstable due to residual noise resulting from the slow convergence in the first step.

\section{Conclusion and perspectives}\label{sec:conclusion}

In this paper, we enhanced the 2D line super-resolution technique initiated in \cite{polisano2019Convex}, by reducing line detection to a spike search in the Radon space through two types of kernels: one characterizing lines convolved with a Gaussian PSF, and the other characterizing ridges in the spectrogram of a superposition of linear chirps. When applied to these kernels, the use of the SFW algorithm provides a very good quality of line estimation, even in the presence of strong degradations (blur, noise, interference). As a result, several perspectives arise: one could combine the two approaches by first denoising the image via atomic norm minimization \cite{polisano2019Convex} (resulting in a significant noise reduction in the early iterations), and then applying SFW to the denoised image to further improve the quality of the estimation. Another perspective would be to apply variational methods on the spectrogram to separate interference from geometric components, and then apply SFW on the latter to estimate the ridges. Finally, we plan to use our approach to reconstruct the ridges of more complex signals by locally approximating them with lines and stitching the estimations over these patches.  

\section*{Acknowledgments}

The authors would like to thank Quentin Denoyelle for the discussions regarding the SFW algorithm and its implementation, which served as the basis for this work.

\bibliographystyle{IEEEtran}
\bibliography{references.bib}

\begin{thebibliography}{10}
\providecommand{\url}[1]{#1}
\csname url@samestyle\endcsname
\providecommand{\newblock}{\relax}
\providecommand{\bibinfo}[2]{#2}
\providecommand{\BIBentrySTDinterwordspacing}{\spaceskip=0pt\relax}
\providecommand{\BIBentryALTinterwordstretchfactor}{4}
\providecommand{\BIBentryALTinterwordspacing}{\spaceskip=\fontdimen2\font plus
\BIBentryALTinterwordstretchfactor\fontdimen3\font minus
  \fontdimen4\font\relax}
\providecommand{\BIBforeignlanguage}[2]{{%
\expandafter\ifx\csname l@#1\endcsname\relax
\typeout{** WARNING: IEEEtran.bst: No hyphenation pattern has been}%
\typeout{** loaded for the language `#1'. Using the pattern for}%
\typeout{** the default language instead.}%
\else
\language=\csname l@#1\endcsname
\fi
#2}}
\providecommand{\BIBdecl}{\relax}
\BIBdecl

\bibitem{boyd2017alternating}
N.~Boyd, G.~Schiebinger, and B.~Recht, ``The alternating descent conditional
  gradient method for sparse inverse problems,'' \emph{SIAM Journal on
  Optimization}, vol.~27, no.~2, pp. 616--639, 2017.

\bibitem{denoyelle2019sliding}
Q.~Denoyelle, V.~Duval, G.~Peyr{\'e}, and E.~Soubies, ``The sliding
  frank--wolfe algorithm and its application to super-resolution microscopy,''
  \emph{Inverse Problems}, vol.~36, no.~1, p. 014001, 2019.

\bibitem{pham2021optical}
T.-A. Pham, E.~Soubies, F.~Soulez, and M.~Unser, ``Optical diffraction
  tomography from single-molecule localization microscopy,'' \emph{Optics
  Communications}, vol. 499, p. 127290, 2021.

\bibitem{candes2014towards}
E.~J. Cand{\`e}s and C.~Fernandez-Granda, ``Towards a mathematical theory of
  super-resolution,'' \emph{Communications on pure and applied Mathematics},
  vol.~67, no.~6, pp. 906--956, 2014.

\bibitem{bhaskar2013atomic}
B.~N. Bhaskar, G.~Tang, and B.~Recht, ``Atomic norm denoising with applications
  to line spectral estimation,'' \emph{IEEE Transactions on Signal Processing},
  vol.~61, no.~23, pp. 5987--5999, 2013.

\bibitem{condat2020atomic}
L.~Condat, ``Atomic norm minimization for decomposition into complex
  exponentials and optimal transport in fourier domain,'' \emph{Journal of
  Approximation Theory}, vol. 258, p. 105456, 2020.

\bibitem{polisano2019Convex}
K.~Polisano, L.~Condat, M.~Clausel, and V.~Perrier, ``A {{Convex Approach}} to
  {{Superresolution}} and {{Regularization}} of {{Lines}} in {{Images}},''
  \emph{SIAM Journal on Imaging Sciences}, vol.~12, no.~1, pp. 211--258, 2019.

\bibitem{polisano2016Convex}
------, ``Convex super-resolution detection of lines in images,'' in \emph{2016
  24th {{European Signal Processing Conference}} ({{EUSIPCO}})}, 2016, pp.
  336--340.

\bibitem{vandenberghe1996semidefinite}
L.~Vandenberghe and S.~Boyd, ``Semidefinite programming,'' \emph{SIAM review},
  vol.~38, no.~1, pp. 49--95, 1996.

\bibitem{condat2013}
L.~Condat, ``A primal--dual splitting method for convex optimization involving
  {L}ipschitzian, proximable and linear composite terms,'' \emph{Journal of
  Optimization Theory and Applications}, vol. 158, no.~2, pp. 460--479, 2013.

\bibitem{rahman1987total}
M.~Rahman and K.-B. Yu, ``Total least squares approach for frequency estimation
  using linear prediction,'' \emph{IEEE Transactions on Acoustics, Speech, and
  Signal Processing}, vol.~35, no.~10, pp. 1440--1454, 1987.

\bibitem{meignen2022one}
S.~Meignen, N.~Laurent, and T.~Oberlin, ``One or two ridges? an exact mode
  separation condition for the gabor transform,'' \emph{IEEE Signal Processing
  Letters}, vol.~29, pp. 2507--2511, 2022.

\bibitem{pham2018novel}
D.-H. Pham and S.~Meignen, ``A novel thresholding technique for the denoising
  of multicomponent signals,'' in \emph{2018 IEEE International Conference on
  Acoustics, Speech and Signal Processing (ICASSP)}.\hskip 1em plus 0.5em minus
  0.4em\relax IEEE, 2018, pp. 4004--4008.

\end{thebibliography}

\end{document}